\documentclass[fleqn,10pt]{wlscirep}
\usepackage[utf8]{inputenc}
\usepackage[T1]{fontenc}
\usepackage[misc]{ifsym}
\usepackage{lineno}
\usepackage{multirow}
\usepackage{xcolor}
\usepackage{soul}
\usepackage{colortbl}
\usepackage{tabularray}
\usepackage{tabularx}
\usepackage{lipsum}
\usepackage{hyperref}
\usepackage{array}
\usepackage{makecell}
\usepackage{float}

\usepackage{algorithm}
\usepackage{graphicx}
\usepackage{textcomp}
\usepackage[singlelinecheck=off]{caption}

\usepackage{amsmath,amssymb,amsfonts}
\usepackage{algpseudocode}
\usepackage{cite}

\usepackage{hyperref}

\algnewcommand{\LeftComment}[1]{\Statex \(\triangleright\) #1}

\def\BibTeX{{\rm B\kern-.05em{\sc i\kern-.025em b}\kern-.08em
    T\kern-.1667em\lower.7ex\hbox{E}\kern-.125emX}}

\makeatletter
\def\hlinewd#1{%
\noalign{\ifnum0=`}\fi\hrule \@height #1 %
\futurelet\reserved@a\@xhline}
\makeatother

\title{A Pan-cancer Classification Model using Multi-view Feature Selection Method and Ensemble Classifier}

\author[ \Letter]{Tareque Mohmud Chowdhury}
\author{Farzana Tabassum}
\author{Sabrina Islam}
\author{Abu Raihan Mostofa Kamal}

\affil{Islamic University of Technology, Gazipur, 1704, Dhaka, Bangladesh}
\affil[ \Letter ]{corresponding author: Tareque Mohmud Chowdhury (tareque@iut-dhaka.edu)}
\begin{abstract}
\textit{Introduction:} Accurately identifying cancer samples is crucial for precise diagnosis and effective patient treatment. Traditional methods falter with high-dimensional and high feature-to-sample count ratios, which are critical for classifying cancer samples. This study aims to develop a novel feature selection framework specifically for transcriptome data and propose two ensemble classifiers. \\

\textit{Methods:} For feature selection, we partition the transcriptome dataset vertically based on feature types. Then apply the Boruta feature selection process on each of the partitions, combine the results, and apply Boruta again on the combined result. We repeat the process with different parameters of Boruta and prepare the final feature set. Finally, we constructed two ensemble ML models based on LR, SVM and XGBoost classifiers with max voting and averaging probability approach. We used 10-fold cross-validation to ensure robust and reliable classification performance. \\

\textit{Results:} With 97.11\% accuracy and 0.9996 AUC value, our approach performs better compared to existing state-of-the-art methods to classify 33 types of cancers. A set of 12 types of cancer is traditionally challenging to differentiate between each other due to their similarity in tissue of origin. Our method accurately identifies over 90\% of samples from these 12 types of cancers, which outperforms all known methods presented in existing literature. The gene set enrichment analysis reveals that our framework's selected features have enriched the pathways highly related to cancers. \\ 

\textit{Conclusions:} This study develops a feature selection framework to select features highly related to cancer development and leads to identifying different types of cancer samples with higher accuracy.

\keywordname{Pan-cancer, Cancer-classification, RNA-seq, TCGA, Feature-Selection, Feature-Partition, Ensemble-Method, Machine-Learning}

\end{abstract}

\begin{document}

\flushbottom
\maketitle

\thispagestyle{empty}

% \noindent Please note: Abbreviations should be introduced at the first mention in the main text – no abbreviations lists or tables should be included. Structure of the main text is provided below.

\section{Introduction}
\label{sec:introduction}
\par Cancer, a heterogeneous group of diseases characterized by uncontrolled cell growth, poses a significant global health burden. The International Agency for Research on Cancer (IARC) has provided statistical records indicating that in 2022, there were approximately 20 million new cancer cases and 9.7 million cancer-related deaths\cite{cancertoday}, or 1 in 6 deaths worldwide, and it is the second leading cause of mortality globally, after cardiovascular disease\cite{who_cancer}. Traditional cancer classification and identification systems, which mostly rely on the site of origin and histological data\cite{TNM}, often fall short in capturing the underlying molecular differences for different cancer types. This limitation drives the necessity for a more precise classification method to enhance diagnostic accuracy, prognostic evaluation, and therapeutic targeting. Recent advancements in high-throughput technologies\cite{HTS}, particularly in the realm of gene expression profiling\cite{GEP}, have opened new avenues for understanding the complex molecular underpinnings of various cancer types. RNA-Seq gene expression data, quantifying the levels of RNA transcripts in a cell or tissue, provides a comprehensive snapshot of cellular activity at the macromolecular level. By analyzing these data, researchers can identify distinct molecular patterns that signify different cancer types. However, gene expression profiling produces a high-dimensional dataset that is challenging for effective analysis and interpretation using conventional techniques. Machine learning (ML) algorithms are incredibly powerful for analyzing such large amounts of data. MLs have been widely applied to evaluate gene expression data in a variety of fields, including disease classification, gene identification, and examining the features of disease progression\cite{NGS_uses1,NGS_uses2,NGS_uses3}, etc.

\par Cancer classification is a critical step in the diagnostic process, which benefits immensely from accurate and robust classification models. Traditional methods often struggle to handle the complexity and heterogeneity of cancer-related, high-dimensional gene expression datasets. In contrast, machine learning techniques have demonstrated considerable success in this domain, offering sophisticated algorithms that can discern intricate patterns in large datasets. Most tumor (cancer) studies focused on the same tumor type, including normal and tumor sample identification, differential gene expression, regulatory gene network identification, sub-type classification, etc. However, the molecular level of genomic data must reflect tumor heterogeneity between different tumor types. Therefore, to understand and capture the similarity and dissimilarity between different tumors, TCGA\cite{TCGA} launched the Pan-Cancer Analysis Project\cite{TCGA_Pan_Cancer} in 2013. Thereafter, pan-cancer analysis is increasing gradually, and researchers are finding tumor related genes to accurately classify and predict the type of tumor. Many studies have already developed different machine learning (ML) models to classify tumor samples with higher accuracy. For example, Wang et al.\cite{review7} proposed a GA/KNN model to classify 10267 samples of 33 tumor types and achieved an accuracy of 96.81\%. Each sample of the dataset contains 20531 features, which were substantially reduced using the mutual information (MI) based feature selection method.

\par The performance of ML models to analyze high-dimensional gene expression datasets heavily depends on the selection of relevant informative features, which is not a trivial task. Effective feature selection can significantly enhance the model's accuracy, reduce overfitting, and improve the interpretability of the results\cite{feature1}, whereas irrelevant or redundant features can obscure meaningful biological insights and degrade classifier performance. There are numerous techniques for feature selection, ranging from filter\cite{feature_filter} and wrapper\cite{feature_wrapper} methods to embedded\cite{feature_embedded} methods. Each of these approaches has its strengths and limitations, and their effectiveness can vary depending on the specific characteristics of the data. Given the limitations of single-method approaches\cite{feature_limitations}, there is a growing interest in the development of ensemble methods\cite{feature_ensemble1, feature_ensemble2} for feature selection. Ensemble methods combine multiple feature selection strategies to achieve more robust, stable, and accurate selection. One ensemble approach for feature selection, multi-view or feature partitioning, was applied to high-dimensional datasets in several studies\cite{feature_partition1,feature_partition2,feature_partition3,feature_partition4}. These methods can leverage the complementary strengths of different approaches, thereby mitigating their individual weaknesses. In the context of cancer classification, an ensemble feature selection approach can help in identifying a more predictive set of features, thereby enhancing the performance of the classification model.

\par This study proposed a feature selection approach based on feature partitioning for high-dimensional gene expression dataset. Also, two ensemble machine learning models were proposed based on selected feature sets to classify 33 types of cancer. We evaluated the models using standard metrics such as accuracy, f1-score, kappa score, precision, and recall. The proposed methods, average voting parallel ensemble (avEns) and max voting parallel ensemble (mvEns), achieved overall test accuracy of 97.11\% and 96.88\%. We can summarize this manuscript's main contributions as follows: 

\begin{enumerate}
%\item Pan-cancer transcriptomic data were collected from the TCGA repository for 33 types of tumors. After preprocessing, two datasets have been prepared. Dataset TCGA33 with 33 classes (33 types of tumor samples) and dataset TCGA34N with 34 classes (33 types of tumor and 1 normal sample).

\item To select informative features, we applied the feature space partitioning approach to the high-dimensional gene expression dataset. During the process, we partition the feature space by feature types, which had not been proposed previously to the best of our knowledge. To carry out further analysis, we identify 12 ranked feature sets and select the set with the best classification performance.

\item Extensive experiments and analysis were performed using proposed average-voting parallel ensemble (avEns) and max-voting parallel ensemble (mvEns) classifier models along with benchmark classifiers, and the results were compared with existing work through evaluation metrics. All experiments were performed on the basis of 10-fold cross-validation to assess the model's adaptability to a wider context. In addition, it helps to reduce overfitting, which is a common phenomenon in machine learning.
\end{enumerate}

\par The remainder of this paper is organized as follows: The literature review section provides a review of the literature and outlines the research gap. The Materials and Methods section covers dataset preparation, feature selection and classification models, and performance evaluation metrics. The Results and Discussion section presents the results achieved by the models, discusses the performance of the models, and compares them with existing literature. Finally, the concluding section summarizes the key findings and insights.

\section{Literature Review}
Researchers have developed numerous machine learning techniques for the analysis of pan-cancer samples. We conducted a literature review of the latest studies in this field to identify tumor samples using TCGA RNA-Seq gene expression data. During the  review, we focus on studies that classified at least 21 TCGA tumor types and performed multi-fold cross-validation to reduce variability. The summary of the related works on the TCGA RNA-Seq dataset for tumor classification is presented in \autoref{tab:literature_review}.

\begin{table*}[ht]
\caption{Summary of tools, technologies, and models used for TCGA tumor samples classification.}
\label{tab:literature_review}
\begin{tabular}{|l|l|l|l|l|l|l|} \hline
\textbf{Research work} & \textbf{Classes} & \textbf{Feature space} & \textbf{Initial feature selection}& \textbf{Model} & \textbf{Performance} \\ \hline
Wang et al.\cite{review7} & 33 tumors & mRNA & MI & \makecell[l]{DL: DenseNet\\(2D-CNN)} & Accuracy: 96.81 \% \\ \hline
Lyu et al.\cite{review10} & 33 tumors & mRNA & Variance Thresold & DL: 2D CNN & Accuracy: 95.59\% \\ \hline
Hsu et al.\cite{review9} & 33 tumors & mRNA & Variance Thresold & ML: linear SVM & Accuracy: 95.8\% \\ \hline
Li et al.\cite{review1} & 31 tumors & mRNA & - & ML: GA/KNN & Accuracy: 90\% \\ \hline
Guia et al.\cite{review2} & 33 tumors & mRNA & - & DL: 2D CNN & Accuracy: 95.65\% \\ \hline
\makecell[l]{Almuayqil \\et al.\cite{review3}} & \makecell[l]{33 tumors\\+1 normal} & mRNA & Lasso, RF, Chi-square & DL: D-CNN & \makecell[l]{Accuracy: 96.16\%,\\Precision:  94.11\%,\\Recall: 94.26\%,\\F1-score: 94.14\%} \\ \hline
\makecell[l]{Mostavi1 \\et al.\cite{review4}} & \makecell[l]{33 tumors\\+1 normal} & transcriptome & \makecell[l]{mean and \\standard deviation} & \makecell[l]{DL: 1D-CNN,\\DL: 2D-Vanilla-CNN,\\DL: 2D-Hybrid-CNN} & \makecell[l]{Accuracy: 95.50\%, \\ Accuracy: 94.87\%, \\ Accuracy: 95.7\%} \\ \hline
Ramirez et al.\cite{review5} & \makecell[l]{33 tumors\\+1 normal} & transcriptome & \makecell[l]{mean and \\standard deviation} & DL: GCNN & Accuracy: 92.76\% \\ \hline
Hossain et al.\cite{review8} & 21 tumors & transcriptome & DEG (75 features/class) & ML: PC-RMTL & Accuracy: 96.07\% \\ \hline
\end{tabular}
\end{table*}

\par Wang et al. \cite{review7} proposed a method to convert gene expression data into 2D images and apply 2D-CNN based DenseNet to classify 33 types of tumor samples and achieved an outstanding accuracy of 96.81\%. Before training the classifier, they reduce the high-dimensional feature set using Mutual Information (MI)\cite{mutual_information} metrics based on information entropy\cite{information_entropy1,information_entropy2}. This approach shows less than 90\% accuracies (0\%, 75\%, 83\%, 85\%, 89\%) towards READ, CHOL, UCS, ESCA, and KICH samples. Another 2D-CNN model proposed by Lyu et al. \cite{review10} achieves an accuracy of 95.59\% for 33 tumor types, showing weakness to identify exactly the same set of tumors, i.e., READ, CHOL, ESCA, UCS, and KICH with reported accuracies of 35\%, 56\%, 77\%, 81\%, and 87\%, respectively. This study uses variance threshold approach for initial feature reduction. Hsu et al. \cite{review9} focused on the performance of five machine learning (ML) models to classify TCGA 33 types of tumor samples. They implemented decision tree (DT), k nearest neighbor (KNN), artificial neural network (ANN), linear support vector machine (linear SVM), and polynomial support vector machine (poly SVM) and achieved the highest accuracy of 95.80\% among them for linear SVM. This study used variance threshold method to reduce feature dimension before model training. This study did not report class-wise classification performance. Li et al. \cite{review1} developed a k-nearest neighbors method with an embedded genetic algorithm (GA/KNN) to classify 31 types of tumor samples. This study did not include ESCA and STAD tumors for their analysis. Moreover, they did not integrate any model pre-training feature reduction strategy. This model achieved overall 90.00\% accuracy on the mRNA gene expression dataset and 0\%, 62\%, and accuracy reported for READ, UCS, and CHOL samples. In \cite{review2}, de Guia et al. proposed a 2D-CNN based deep learning model and claimed 95.65\% accuracy for classifying TCGA 33 tumor samples by analyzing mRNA expression data. For 5 tumor classes (READ, CHOL, ESCA, UCS, and KICH), they achieve poor classification accuracy (35\%, 56\%, 77\%, 81\%, and 87\%, respectively). However, the class-wise performance in their paper did not include STAD class. Moreover, they did not use any feature selection process to reduce feature space before training the model. Almuayqil et al. \cite{review3} developed a 2D-CNN deep learning model in their study to identify 34 tumor classes (33 tumors and 1 normal) extracted from the TCGA repository. They use mRNA expression data and apply three standard feature selection techniques, namely Lasso Regression, Random-Forest, and Chi-square, to reduce the hi-dimensional feature space. This study reported overall 96.16\% accuracy, 94.11\% precision, 94.26\% recall, and 94.14\% F1-score performance. Also, tumor classes CHOL, READ, ESCA, COAD, and KICH have shown accuracy of 57\%, 61\%, 85\%, 88\%, and 88\%, respectively. In \cite{review4}, Mostavi et al. implemented three CNN models: 1D-CNN, 2D-Hybrid-CNN, and 2D-Vanilla-CNN and achieved accuracies of 95.50\%, 94.87\%, and 95.70\%, respectively, for classifying 34 tumor classes downloaded from the TCGA repository. Before training the model, they reduce feature space by removing features with low information burden (mean < 0.5 or standard deviation < 0.8) across all samples. This paper did not report class-wise classification performance on the test dataset. Ramirez et al. \cite{review5} proposed three variants of Graph Convolutional Neural Networks (GCNN) to classify 34 tumor classes and achieved accuracies ranging from 87.75\% to 92.76\%. After downloading transcriptome data from the TCGA repository, they apply the $log_2(FPKM+1)$ transformation and keep informative features with a mean > 0.5 and a standard deviation > 0.8 across all samples. However, this study did not report class-wise performance on the test dataset in their paper. In a study\cite{review8}, Hossain et al. extended a regularized multi-task learning (RMTL) \cite{rmtl} model for pan-cancer classification (PC-RMTL) and achieved an excellent accuracy of 96.07\% to classify 21 tumor types and associated normal samples using RNA-Seq data collected from TCGA repository. For initial feature selection, they have collected the top 75 differentially expressed genes (DEGs) using the DESeq2 package in R for each of the 21 cancer types by comparing the tumor and adjacent normal samples. Then they pick unique features from the 21 DEG sets, which leads to 1055 features. Hoadley et al.\cite{CellofOrigin} analyzed 33-type cancer samples from TCGA repository and suggested that the cell-of-origin patterns dominate the classification of cancer samples. CHOL and LIHC are both common primary liver cancer\cite{CHOL_LIHC}. PAAD originates in the pancreas attached to the liver. COAD and READ are both originated from similar tissue and combined to be referred to as colorectal cancer(CRC)\cite{READ_COAD}. Both uterine carcinosarcoma (UCS) and uterine corpus endometrial carcinoma (UCEC) tumors originate around the uterus. ESCA and STAD are misclassified with each other, whereas they both originated in the digestive tract. Previously, Quin et al.\cite{ESCA_STAD} have shown that STAD and ESCA have some common molecular biomarkers. KICH, KIRC, and KIRP all originate in kidney tissue\cite{KICH_KIRC_KIRP}.

% research articles \cite{review1,review2,review3,review4,review5,review8,review10}

\par Class-wise accuracy provided in the above articles suggests that tumors that are classified with low accuracy are mostly misclassified as tumors within the same tissue of origin. For example, the tumor sets \{CHOL, LIHC, PAAD\} originated from the same tissue, and they mostly misclassified within themselves. The same is true for sets \{READ, COAD\}, \{ESCA, STAD\}, \{KICH, KIRC, KIRP\}, and \{UCEC, UCS\}. Allmost all of the tumor classification methods reported so far have shown poor average accuracy (<90\%) to classify or identify these twelve tumors, which is shown in \autoref{tab:compare_critial}. From the review, it can be concluded that existing feature selection and classification methods are not strong enough to classify these 12 tumor types with acceptable accuracy.

\section{Material and Methods}
Our study consists of three main sections: 1) Dataset preparation, 2) Feature selection and 2) Classification models. In the dataset preparation stage, we download data for various tumors from TCGA repository, filter out unnecessary information, merge them to build a single dataset followed by normalization steps. In the second stage, We propose a pipeline to identify informative features crucial for the classification. We evaluated the performance of different state-of-the art feature selection techniques and compare those with our proposed feature selection method. Finally, we employ two ensemble models to classify tumors accurately and compare it's performance with different existing standard classifiers. 

\begin{table*}[!htbp]
\caption{Summary of pan-cancer data downloaded from TCGA repository.}
\centering
\label{tab:tcgadata}
    \begin{tabular}{|l|c|c|c|c|c|}
    \hline
        \multicolumn{3}{|c|}{\textbf{Tumor Specification}} & \multicolumn{3}{|c|}{\textbf{Samples}} \\ \hline
        Tumor Name & Abbreviation & Project Name & Total  & Normal & Tumor \\ \hline
        Adrenocortical carcinoma & ACC & TCGA-ACC & 79 & 0 & 79 \\ \hline
        Bladder Urothelial Carcinoma & BLCA & TCGA-BLCA & 431 & 19 & 412 \\ \hline
        Breast invasive carcinoma & BRCA & TCGA-BRCA & 1231 & 113 & 1111 \\ \hline
        Cervical squamous cell carcinoma & CESC & TCGA-CESC & 309 & 3 & 304 \\ \hline
        Cholangiocarcinoma & CHOL & TCGA-CHOL & 44 & 9 & 35 \\ \hline
        Colon adenocarcinoma & COAD & TCGA-COAD & 524 & 41 & 481 \\ \hline
        Lymphoid Neoplasm Diffuse B-cell Lymphoma & DLBC & TCGA-DLBC & 48 & 0 & 48 \\ \hline
        Esophageal carcinoma & ESCA & TCGA-ESCA & 198 & 13 & 184 \\ \hline
        Glioblastoma multiforme & GBM & TCGA-GBM & 175 & 5 & 157 \\ \hline
        Head and Neck squamous cell carcinoma & HNSC & TCGA-HNSC & 566 & 44 & 520 \\ \hline
        Kidney Chromophobe & KICH & TCGA-KICH & 91 & 25 & 66  \\ \hline
        Kidney renal clear cell carcinoma & KIRC & TCGA-KIRC & 614 & 72 & 541 \\ \hline
        Kidney renal papillary cell carcinoma & KIRP & TCGA-KIRP & 323 & 32 & 290 \\ \hline
        Acute Myeloid Leukemia & LAML & TCGA-LAML & 151 & 0 & 151 \\ \hline
        Brain Lower Grade Glioma & LGG & TCGA-LGG & 534 & 0 & 516 \\ \hline
        Liver hepatocellular carcinoma & LIHC & TCGA-LIHC & 424 & 50 & 371 \\ \hline
        Lung adenocarcinoma & LUAD & TCGA-LUAD & 600 & 59 & 539 \\ \hline
        Lung squamous cell carcinoma & LUSC & TCGA-LUSC & 553 & 51 & 502 \\ \hline
        Mesothelioma & MESO & TCGA-MESO & 87 & 0 & 87 \\ \hline
        Ovarian serous cystadenocarcinoma & OV & TCGA-OV & 429 & 0 & 421 \\ \hline
        Pancreatic adenocarcinoma & PAAD & TCGA-PAAD & 183 & 4 & 178 \\ \hline
        Pheochromocytoma and Paraganglioma & PCPG & TCGA-PCPG & 187 & 3 & 179 \\ \hline
        Prostate adenocarcinoma & PRAD & TCGA-PRAD & 554 & 52 & 501 \\ \hline
        Rectum adenocarcinoma & READ & TCGA-READ & 177 & 10 & 166 \\ \hline
        Sarcoma & SARC & TCGA-SARC & 265 & 2 & 259 \\ \hline
        Skin Cutaneous Melanoma & SKCM & TCGA-SKCM & 473 & 1 & 103 \\ \hline
        Stomach adenocarcinoma & STAD & TCGA-STAD & 448 & 36 & 412 \\ \hline
        Testicular Germ Cell Tumors & TGCT & TCGA-TGCT & 156 & 0 & 150 \\ \hline
        Thyroid carcinoma & THCA & TCGA-THCA & 572 & 59 & 505  \\ \hline
        Thymoma & THYM & TCGA-THYM & 122 & 2 & 120 \\ \hline
        Uterine Corpus Endometrial Carcinoma & UCEC & TCGA-UCEC & 589 & 35 & 553 \\ \hline
        Uterine Carcinosarcoma & UCS & TCGA-UCS & 57 & 0 & 57 \\ \hline
        Uveal Melanoma & UVM & TCGA-UVM & 80 & 0 & 80  \\ \hline
        \multicolumn{3}{|l|}{Total samples}  & 10818 & 740 & 10078 \\ \hline
    \end{tabular}
\end{table*}

\subsection{Dataset Preparation}
\par The gene expression dataset for 33 different cancer types is downloaded from the Cancer Genome Atlas (TCGA) project\cite{TCGA} of Genomic Data Commons(GDC)\cite{GDC} using TCGAbiolinks library\cite{TCGABiolinks} in the R studio environment. There are a number of parameters (project, legacy, data.category, data.type, workflow.type) that we should set to pull our desired data from the TCGA repository using the GDCquery function. The project argument refers to the project name, which represents the cancer type. The legacy argument can be either true or false. When false, the query should look into the harmonized data repository when fetching the data. Harmonized data\cite{Harmonized} is a compilation of unprocessed data from several sources that has been standardized to allow for accurate comparisons between these sources. The parameter data.category refers to the category of data in the specific project. In our case, we set the data category to “Transcriptome Profiling." The data.type argument refers to the data type that we can use for filtering the files to download. In our case, we set it to “gene expression quantification." Genes can be quantified by counting the number of reads that map to each gene using RNA-Seq\cite{RNASeq}. Finally, we set workflow.type to “STAR - Counts” which represents reads that are mapped to the reference genome using the STAR algorithm\cite{STAR}. The downloaded gene expression data for each project, i.e., cancer type, is transformed into a SummerisedExperiment object\cite{SE} with six matrices, namely fpkm\_unstrand, fpkm\_uq\_unstrand, stranded\_first, stranded\_second, tpm\_unstrand, and unstranded. In our study, we have used fpkm\_unstrand matrix, where the read counts are normalized by the fpkm method. The columns in this matrix represent the samples (or cases), and the rows represent ensemble IDs for 60,660 genes. After downloading the gene data, we have 33 SummerizedExperiment objects for 33 cancer types, totaling 11,274 samples.

\subsubsection*{Data preproccessing}
\par We extracted gene expression data from the fpkm\_unstranded assay of downloaded SummerizedExperiment objects, where sample types are either “Solid Tissue Normal” or “Primary Tumor." For LAML cancer type, we use the sample type “Primary Blood Derived Cancer - Peripheral Blood” instead of “Primay Tumor." We extracted a total of 10,078 samples across 33 tissue types with 60,660 features, which has been shown in \autoref{tab:tcgadata}. Then, transpose the dataframe matrix so that features are in columns and samples are in rows. To reduce dimensionality we remove features with very low expression value (average value < 0.05). This step reduces feature counts to 36,017. As the normalized expression values in the prepared dataset are sparse, with the highest value being 549,980 and the lowest value being 0 with a mean of 9.28 and a standard deviation of 214.87, we applied binary logarithm ($log_{2}$) to the data values. The logarithm transforms the values into a more linear form and placed on a comparable scale\cite{logtransform}. To avoid NaN errors, we added 1 to each value before applying the log. We name the prepared dataset TCGA33Tumors\cite{tcga33tumors}, which is publicly available in Harvard Dataverse.

\subsubsection*{Dataset Partitioning}
The prepared dataset has a high-dimensional feature set with a total of 36,017 features (genes) and 10,078 samples. We partition the dataset vertically (column-wise) based on the feature types, i.e., messenger RNA (mRNA), micro RNA (miRNA), long non-coding RNA (lncRNA), and other RNA. The partitioned datasets contain 17,695, 639, 8,961, and 8,758 features (columns), respectively. Partition wise minimum, maximum, mean and standard deviation of the values are shown in \autoref{tab:featuregroup}. This table presents a glimpse of statistical differences between values of dataset partitions. 

\begin{table}[]
\caption{Differences in mean and standard deviation of values between dataset partitions after log transformation.}
\label{tab:featuregroup}
\begin{tabular}{|l|ccccc|}
\hline
                   & \multicolumn{5}{c|}{\textbf{Group wise statistics after log transformation }}                                                                                       \\ \hline
\textbf{Metrics}   & \multicolumn{1}{l|}{\textbf{All}} & \multicolumn{1}{l|}{\textbf{mRNA}} & \multicolumn{1}{c|}{\textbf{miRNA}} & \multicolumn{1}{c|}{\textbf{lncRNA}} & \textbf{othRNA} \\ \hline
Frequency          & \multicolumn{1}{c|}{36,017} & \multicolumn{1}{c|}{17,659}         & \multicolumn{1}{c|}{639}           & \multicolumn{1}{c|}{8,961}           & 8,758           \\ \hline
Min value          & \multicolumn{1}{c|}{0} & \multicolumn{1}{c|}{0}             & \multicolumn{1}{c|}{0}               & \multicolumn{1}{c|}{0}              & 0               \\ \hline
Max value          & \multicolumn{1}{c|}{19.07} & \multicolumn{1}{c|}{17.20}      & \multicolumn{1}{c|}{12.08}        & \multicolumn{1}{c|}{13.84}       & 19.07        \\ \hline
Mean               & \multicolumn{1}{c|}{1.20}  & \multicolumn{1}{c|}{2.10}    & \multicolumn{1}{c|}{0.26}      & \multicolumn{1}{c|}{0.33}        & 0.35        \\ \hline
Sd & \multicolumn{1}{c|}{1.65} & \multicolumn{1}{c|}{1.85}     & \multicolumn{1}{c|}{0.64}        & \multicolumn{1}{c|}{0.59}       & 0.81        \\ \hline
\end{tabular}
\end{table}
%\vspace{-10pt}

\subsection{Feature Selection}
The high-dimensional gene expression dataset, prepared from TCGA gene expression data, consists of 36,017 features. The term 'feature' in this context referenced to genes present in the dataset. In this section,  we studied a number of established feature selection methods widely used in the field of Bioinformatics. Then, we propose a new approach to identify informative features valuable for tumor identification. 

\subsubsection*{Established feature selection methods}
\par Well established state-of-the-art feature selection techniques have been studied\cite{estfcomp} for gene expression data and suggested effective methods which include Univariate Feature Selection (UFS), Mutual Information(MI)\cite{MI}, Least Absolute Shrinkage and Selection Operator (LASSO)\cite{Lasso}, Elastic Net (ENet)\cite{EN}, and Recursive Feature Elimination (RFE)\cite{RFE} based feature selection. 

\par UFS aims at finding the k top features by using univarite statistical test scores. UFS aims to find the top k best features by scoring them using univariate statistical tests, such as the ANOVA F-value\cite{Annova}. Mutual information (MI) from the field of information theory is used information gain for feature selection. The high value of mutual information means a closer connection between the feature and the target indicating feature importance for training the model. LASSO uses regression analysis with variable selection and regularization to improve prediction accuracy and interpretability of the statistical model it produces. This model can be used for top k feature selection. ENet tends to balance accuracy and weight size in a linear model, exploiting L1 and L2 regularization, is a popular choice for feature selection in bioinformatics\cite{L12Popularity1,L12Popularity2}. ENet try to reduce features to non-zero weights in the model and takes k high-scored features. RFE runs a machine learning algorithm capable of scoring features, such as Random Forest (RF)\cite{RF}, several times iteratively and removing the feature with the lowest score until it reaches to the user-specified k features. In addition to these, we also consider Chi-Square (Chi2) test\cite{Chi2}, which is a powerful and widely used feature selection method for categorical data in machine learning. All considered feature selection methods are implemented using the machine learning package scikit-learn. 

%Chi2 score is calculated using Chi2 test between each feature and the target and select the desired number of features with best Chi2 scores.

\subsubsection*{Feature Selection by Boruta}
Boruta\cite{boruta} is a wrapper-type feature selection method based on the Random Forest (RF) Classifier. This algorithm started by doubling the width of the dataset by duplicating the original features and adding them to the right side of the dataset. Then, on the duplicate side, values in columns are shuffled to achieve randomness. These shuffled features are called shadow features. After rebuilding the dataset with shadow features, a RF classifier model is trained on this new feature space to determine the importance of features using Z-score. Z-score is a statistical measurement that describes a value's standard deviation from the mean and can be calculated using data point x, mean u, and standard deviation $\delta$ as: \\
\[
Z = \frac{(x-u)}{\delta}
\]

\begin{figure}[t!]
    \includegraphics[width=14cm]{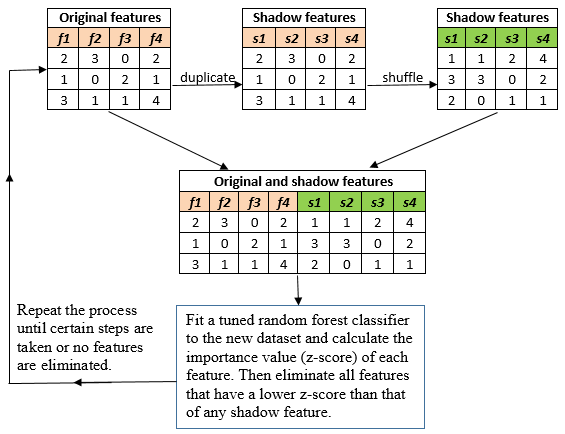}
    \caption{Boruta feature selection process in brief.}
    \label{fig:Boruta}
\end{figure}

\par The Z-score of an original feature is compared with the Z-score of its randomized counterpart. If the compared Z-score is significantly higher, then the feature is marked as \textbf{confirmed} and kept. If the compared Z-score is lower, the feature is marked as \textbf{rejected} and dropped. If the compared z-score is higher but not significantly higher, the feature is marked as \textbf{tentative} and kept for the time being. In the next iteration, another shadow dataset is created using the updated dataset. This process is repeated until all features are \textbf{confirmed/rejected} or until the desired number of iterations are completed. After the final iteration, all \textbf{tentative} features are dropped, leaving only the \textbf{confirmed} features in the dataset. This process is shown in \autoref{fig:Boruta}.

\subsubsection*{Proposed Feature Selection Method}
\begin{figure}[htp]
    \includegraphics[width=17cm]{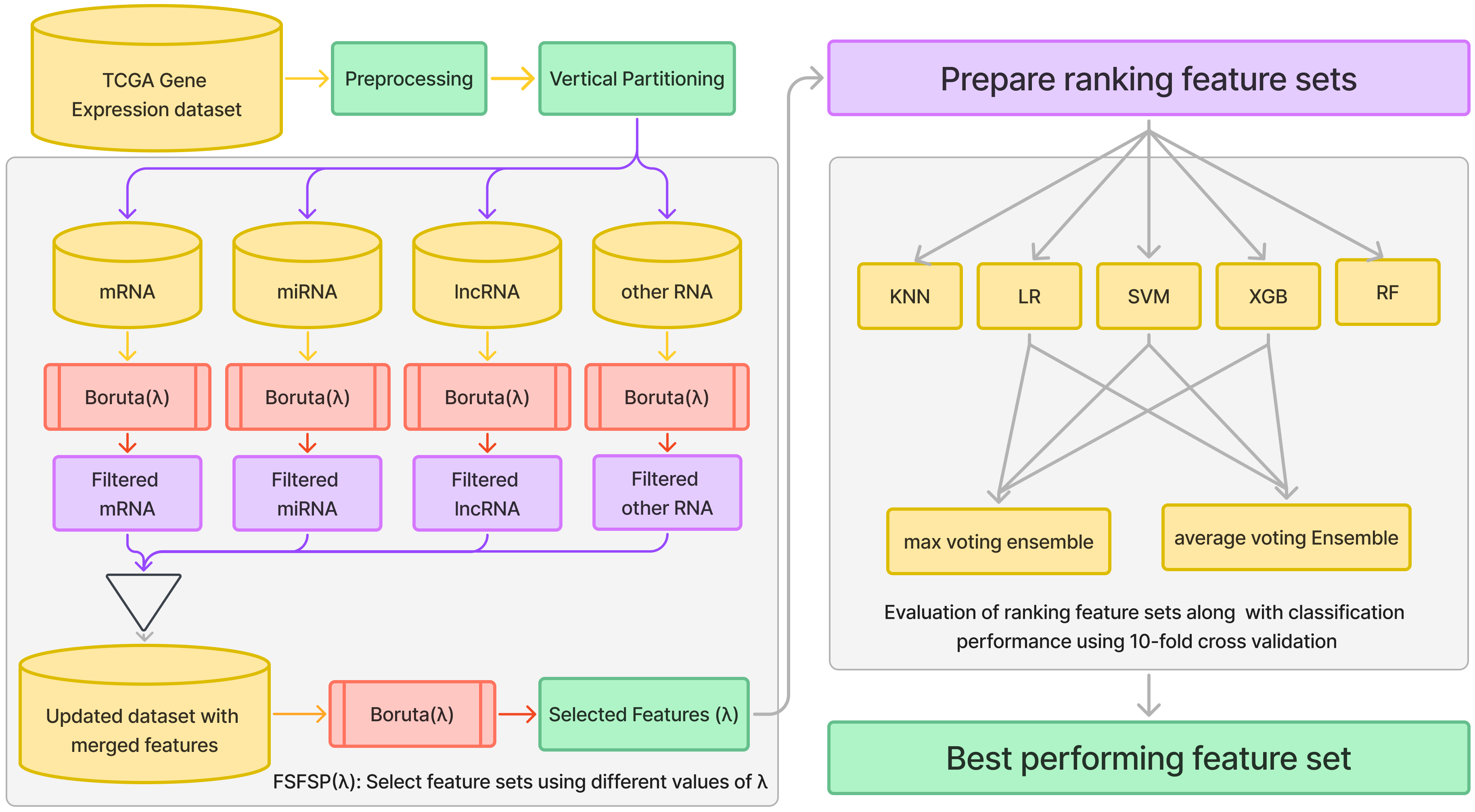}
    \caption{Flow diagram of the proposed feature selection approach and it's performance evaluation.}
    \label{fig:FSFSP}
\end{figure}

To reduce features by eliminating redundant and irrelevant ones from the high-dimensional TCGA dataset, we propose an ensemble feature selection approach, which has been presented in \autoref{fig:FSFSP}. In the proposed method, we partition the dataset vertically based on feature types, then apply the Boruta feature selection algorithm on each partition, merge the selected features resulted from each partition, and update the main dataset by keeping only merged features. Afterwards, we apply the Boruta algorithm again to reduce the feature space.  

\par We tune the hyperparameters for the underlying RF Classifier of Boruta as min\_samples\_split = 6, min\_samples\_leaf= 3, max\_features = 'sqrt', class\_weight2='balanced' and bootstrap2=False. We also set  max\_iter=200 for Boruta and random\_state=1 to reproduce the experiment. Another important hyperparameter max\_depth indicates the number of splits that each Decision Tree in the RF is allowed to make. If max\_depth value is too low, the trained model underfits, and if it is too high, the trained model overfits the data. 
%\cite{RF}

\par As there are no crystal clear boundaries to define the overfitting and underfitting of a model, we propose a method to apply multiple values of hyperparameter \textit{max\_depth} ($\lambda$) across a wider range with a regular interval to find multiple feature sets. Then, we rank each feature based on their presence count in those feature sets. Here, we use 12 different values of $\lambda$, ranging from 5 to 60 with interval 5. We have got 12 overlapped feature sets for different values of $\lambda$. Then, based on the frequencies of each selected feature as a member of these 12 feature sets, we rank the features. $Rank_N$ features is the set of features that are members of at least N out of 12 selected feature sets. The entire process is shown in Algorithm \autoref{alg:feature_rankset}.

\begin{algorithm}[]
\caption{Algorithm to find ranking feature sets using FSFSP method (\autoref{fig:FSFSP}).}
\label{alg:feature_rankset}
\begin{algorithmic}[1]
\State $Function \hspace*{.5em}  RankingFSFSP(dataset)$
\State $FeaturesSelected \gets \{\}$
\State $depth \gets 5$
\While{$depth \leq 60$}
\State $FeatureSets[depth] \gets \newline 
    \hspace*{8.7em} FSFSP(dataset,depth)$
\State $selectedFeatures \gets UNION(selectedFeatures,\newline
    \hspace*{10em} FeatureSets[depth])$    
\State $depth \gets depth + 5$
\EndWhile
%\State $\vspace*{.5em}$
\Statex\LeftComment{Initialize 12 ranking feature sets.}
\For{$k \gets 1$ to $12$}                    
\State {$Rank_k$ $\gets$ {$\{\}$}}
\EndFor
%\State $\vspace*{.5em}$
\Statex\LeftComment{ Progressively update 12 ranking feature sets.}
\For{\textbf{each} feature $\in$ selectedFeatures} 
\State {$Rank_n.insert(feature)$; where feature is a member}
\State {of n out of 12 FeatureSets}
\EndFor
\State $End \hspace*{.5em} Function$
\end{algorithmic}
\end{algorithm}

\subsection{Classifier Models}
We have implemented five cutting edge machine learning classification models to classify pan-cancer transcriptomic data. These models are Logistic Regression(LR)\cite{LR}, Support Vector Machine(SVM)\cite{SVM}, Extreme Gradient Boosting(XGBoost)\cite{XGBoost}, K-Nearest Neighbors(KNN)\cite{KNN}, and Random Forest(RF)\cite{RF}. Then we proposed two parallel ensemble classifier models based on LR, SVM and XGBoost with two different fusion methods, max voting\cite{MaxVoting} and average voting\cite{AverageVoting}. An ensemble classifier is a meta approach to seek better predictive performance by combining (fusion) the predictions from multiple classifier models. Hyperparameter settings of different classifier models are shown in \autoref{tab:hyperparameters}.

\subsubsection*{ML classifiers (Existing) we have used in our study}
\par Logistic Regression (LR), typically used for binary classification, can be extended to multi-class classification via the multinomial logistic regression or softmax regression. This model calculates probabilities of each class based on input features through a softmax function that normalizes computed scores. It's effective for problems with a clear linear boundary between classes. However, its assumption of linear relationships between features and classes limits its use in more complex, nonlinear scenarios. Support Vector Machine (SVM) for multi-class classification involves extending its binary classification approach using methods like one-vs-one or one-vs-all. SVM finds hyperplanes with maximum margin between classes in high-dimensional space, providing robust and accurate classification. It performs well in high-dimensional spaces and is known for its effectiveness in distinguishing classes clearly. However, SVM can be computationally demanding with large datasets and complex multi-class problems. Extreme Gradient Boosting (XGBoost) is a potent multi-class classifier that uses gradient boosting to build sequential decision trees, correcting previous errors and improving accuracy. It employs a one-vs-all approach for multi-class scenarios, optimizing a loss function iteratively. XGBoost is celebrated for its execution speed and model performance, handling large and complex datasets efficiently. It offers scalability and strong predictive capabilities, making it a top choice for challenging classification tasks. K-Nearest Neighbors (KNN) is a straightforward multi-class classifier that assigns classes based on the majority vote from the 'k' nearest training samples to a new point, using distance metrics like Euclidean distance. KNN requires no model training phase, making it simple to implement but computationally intensive with large datasets. It is sensitive to the choice of 'k' and the distance metric, impacting its effectiveness across different scenarios. Random Forest (RF) is an ensemble technique using multiple decision trees to perform multi-class classification. Each tree votes on an input’s class, with the majority vote deciding the final prediction. This method effectively reduces overfitting and enhances accuracy by averaging multiple trees trained on different parts of the dataset. It's robust across different data types and useful for assessing feature importance, making it a versatile choice for complex classification tasks.

\par A crucial step in determining the ideal machine learning parameters is hyper-parameter tuning. We applied grid-search approach to search for hyperparameter values for which the ML models performed better. The hyperparameter values used for methods and models are presented in \autoref{tab:hyperparameters}.

\begin{table}[hb!]
\caption{Adjusted hyperparameters for various models that were employed in the research.}
\label{tab:hyperparameters}
\begin{tabular}{|l|l|}
\hline
\multicolumn{2}{|l|}{\textbf{Boruta Feature Selection Method}}\\ \hline
\textit{Models} & \textit{Hyperparameter settings}\\ \hline
RF & \makecell[l]{min samples split=6, min samples leaf=3,\\ 
max\_features='sqrt', class weight='balanced', \\max depth= 5-60, bootstrap=False} \\ \hline
BorutaPy & \makecell[l]{n estimators='auto', max iter=200} \\ \hline
\multicolumn{2}{|l|}{\textbf{For ML Classifiers}}\\ \hline
\textit{Models} & \textit{Hyperparameter settings} \\ \hline
LR & \makecell[l]{max iter=2000, solver='lbfgs',\\C=20, penalty='l2'} \\ \hline
SVM & \makecell[l]{decision function\_shape='ovo', tol=1e-5,\\ C=1, kernel='linear'} \\ \hline
XGBoost & \makecell[l]{n estimators=1000, max depth= 4,\\ objective="multi:softprob", learning rate=0.1} \\ \hline
KNN & \makecell[l]{n neighbors=7} \\ \hline
RF & \makecell[l]{n estimators=100, max features='sqrt',\\ min samples split=2, min samples leaf=1,\\ criterion='entropy', max depth=12,\\ bootstrap=False} \\ \hline
\end{tabular}
\end{table}

\subsubsection*{Proposed Ensemble Classifiers}
\begin{figure}[htp!]
    \includegraphics[width=10cm]{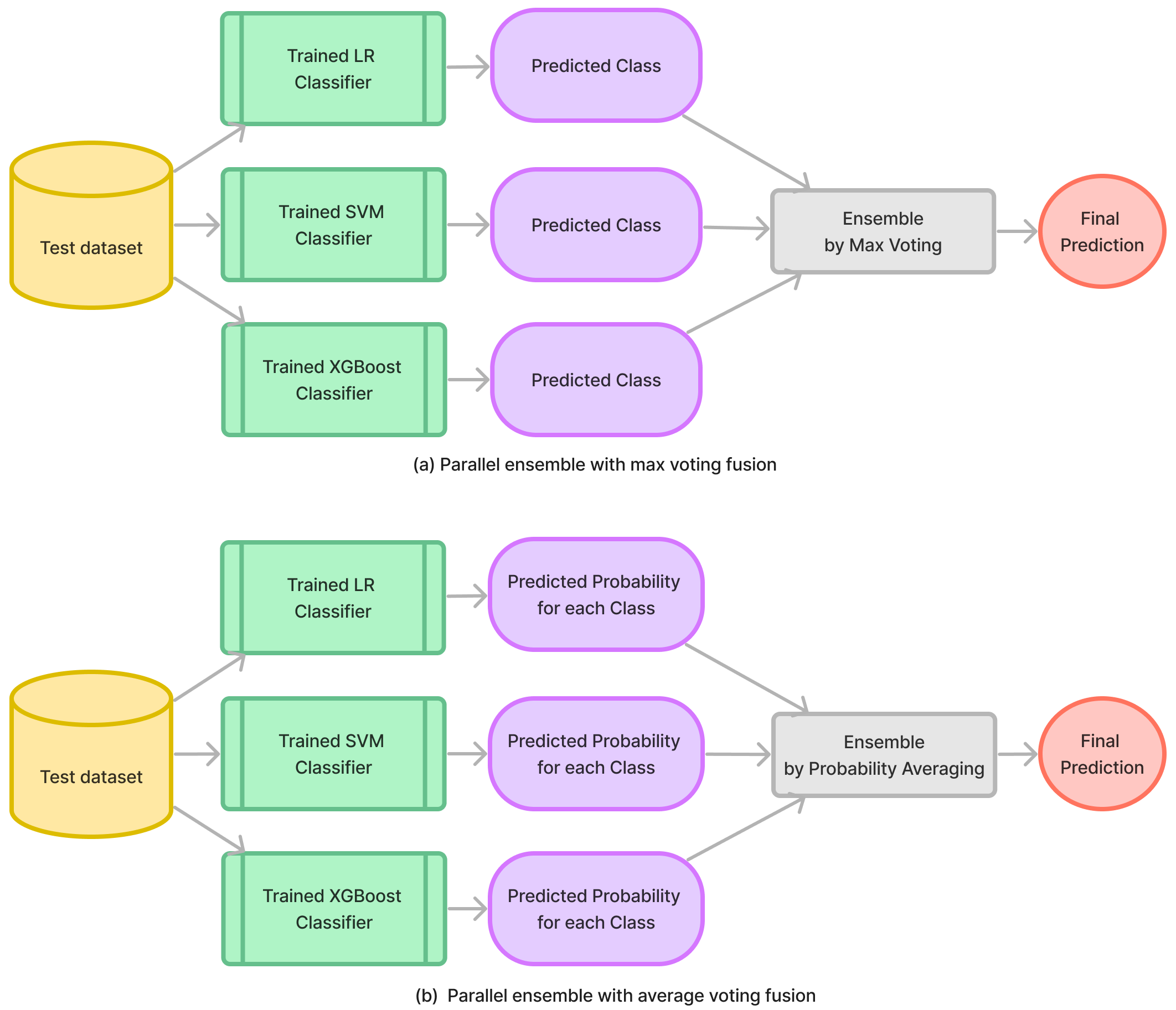}
    \caption{Framework of proposed ensemble models. a) In max voting fusion, the class label with the most votes will be predicted. b) In average voting, the final prediction is derived from an average of the prediction probability derived from component classifiers.}
    \label{fig:Ensemble}
\end{figure}

We proposed two parallel ensemble classifiers in this study: max voting parallel ensemble (mvEns) and averrage voting parallel ensemble (avEns). Both of the models use three baseline classifiers: LR, SVM, and XGBoost.\\ \\
\textbf{Max Voting Parallel Ensemble (mvEns)} \\
The max voting ensemble model uses the max voting method to fuse predicted classes of sub-models for the final prediction, thereby improving the classifier's performance. Max voting entails gathering predictions for every class label and projecting which class label will receive the greatest number of votes. For example, assuming we have three classifiers, C1, C2, and C3, that predict the class for a test sample, [2,2,1] resulted in y = max\_vote[2,2,1] = 2. Hence, the predicted class by the ensemble model is 2. The framework of the model is shown in \textit{part a} of \autoref{fig:Ensemble}.\\ \\
\textbf{Average Voting Parallel Ensemble (avEns)}\\
In an average voting ensemble classifier, we can make a final prediction by extracting predictions from sub-models and averaging them. We compute the average prediction probability by dividing the sum of the prediction probabilities by the total number of predictions. The framework of the model is shown in \textit{part b} of \autoref{fig:Ensemble}. Average voting calculation can be shown using the following equation:

\[y={argmax}_{i}\frac{1}{n}\sum_{j=1}^{m}W_{ij}\] 

where $W_{ij}$ is the predicted probability of the $i^{th}$ class label of the $j^{th}$ classifier.

The average voting method provides a more accurate result because of its predictive power\cite{ensembleModels}. Furthermore, it minimizes overfitting and performs more accurately than max voting.

\subsection{Evaluation Metrics}
We investigated a number of performance metrics to evaluate and compare our proposed method to existing ones. These metrics are namely accuracy, precision, recall, f1-score \cite{metrics}, kappa score\cite{kappa} , and AUC value in ROC analysis. These evaluation metrics are defined below in terms of true positive (TP), true negative (TN), false positive (FP), and false negative (FN) cases.

\subsubsection*{Accuracy (Acc)} Accuracy is defined as the ratio of the number of correctly classified samples to the total number of samples. Accuracy is the metric that gives an initial impression of a classifier's performance. However, accuracy is not sufficient to measure the performance of a classifier, especially if the dataset is imbalanced (which is the case for the datasets we are dealing with). The calculation formula for accuracy is:
\[
Acc = \frac{TP+TN}{TP+FP+FN+TN}
\]
\subsubsection*{Precision (P)} Precision is defined as the ratio of the number of correctly predicted samples to the total number of predicted samples in a class. The calculation formula for precision is:
\[
P = \frac{TP}{TP+FP}
\]
\subsubsection*{Recall (R)} Recall is defined as the ratio of the number of correctly predicted samples to the total number of samples in a class. Recall is also known as sensitivity. The calculation formula for recall is:
\[
R = \frac{TP}{TP+FN}
\]
\subsubsection*{F1 Score (F1)} F1 score is the harmonic mean of precision and recall. The calculation formula for F1-score is:
\[
F1 = \frac{2PR}{P+R}
\]
\subsubsection*{Kappa Score (K)} Kappa score or Cohen Kappa metric is used to evaluate the consistency of classification methods rather than their accuracy. It is particularly useful in scenarios where accuracy is not the only important factor, such as with imbalanced classes.
\[
K = \frac{p_0-p_e}{1-p_e}
\]
\par Where $p_0$ is the empirical probability of agreement on the label assigned to any sample, and $p_e$ is the expected agreement when both annotators (test and prediction) assign labels. $p_e$ is estimated using a per-annotator empirical prior over the class labels.

\subsubsection*{Area under ROC curve (AUC)} A receiver operating characteristic (ROC) curve is a visual metric showing the performance of a classification model at all classification thresholds. This curve can be drawn by plotting the true positive rate (TPR) on the y-axis and the false positive rate (FPR) on the x-axis. TPR is a synonym for recall, and FPR can be calculated as $FP/(FP+TN)$. After drawing the ROC curve, the area under the curve (between the curve and x-axis limiting 0 to 1) is called the AUC\cite{AUC}. Generally, for uniformly distributed dataset, accuracy measure is important for model performance evaluation. However, for imbalanced dataset, ROC AUC value is more important to represent model performance. The ROC\_AUC value for a multiclass model can be evaluated by the roc\_auc\_score() method of the sklearn library with the parameters multi\_class="ovr" and average="weighted" or "macro".

\subsubsection*{Confusion Matrix} A confusion matrix is a table that describes the performance of a classification algorithm. A confusion matrix is a matrix representation that shows and summarizes a classification algorithm's performance. It displays the proportion of accurate and inaccurate predictions for each class. It helps to understand the classes that the model confuses for other classes.

%\subsection*{Hyperparameter Settings}  

\subsection{Experimental Setup} We compose our experiments into three parts. Firstly, we construct twelve overlapping feature sets using the proposed rankingFSFSP feature selection approach and pick the feature set that performs best to classify TCGA tumor gene expression data. Secondly, we implemented two heterogeneous parallel ensemble machine learning models, the max voting ensemble (mvEns) and the average voting ensemble (avEns), based on LR, SVM, and XGB. Finally, we compare the classification performance of our proposed ensemble methods to standard ML models and other methods presented in recent literature related to the classification of the TCGA dataset. We executed all experiments on a Dell EMC PowerEdge R740 server running Ubuntu 20.04 LTS and equipped with 512 GB of RAM using Python v3.12.

\section{Results and Discussion}

\subsection{Feature Selection}
To mitigate the issue of high dimensionality, we remove features with a very low expression value accross the samples (average value < 0.05). Then we apply our proposed feature selection method on the TCGA33Tumors dataset and obtain twelve overlapping ranking feature sets (Rank12, Rank11, Rank10,..., Rank2, Rank1, etc.) with feature counts 1290, 2462, 3515, 4088, 4401, 4699, 5010, 5346, 5010, 5346, 5688, 6076, 6545, and 7170 respectively. The number of features in the ranking feature sets is significantly reduced compared to the total number of 36,017 features in the TCGA33Tumors dataset. Subsequently, we applied five state-of-the-art classifiers to measure the classification accuracy for each of the ranking feature sets. Among the classifiers, LR, SVM and XGBoost showed a higher level of performance for all feature sets. Taking this into account, we explored the effectiveness of two ensemble methods, max voting ensemble (mvEns) and average voting ensemble (avEns), using these three classifiers, aiming to assess the potential enhancements of the classification performance.  

\par To compare the performance of the current state-of-the-art feature selection methods with our proposed one, we select the top 500, 1000, 2000, 3000, 3515, and 4000 features from the trained model using the SelectKBest method. Since the best-performing feature set (rank10) generated by our proposed approach contains 3515 features, we deliberately picked 3515 features (instead of 3500) to make it convenient to compare performance. The classification performance achieved by existing and proposed classifiers using the selected features is shown in \autoref{tab:fslistcompare}. The features selected by the LASSO algorithm provide better classification accuracy compared to other existing feature selection methods. The XGBoost classifier performs better for classification tasks compared to other classifiers for same set of features. However, the features selected by the proposed FSFSP approach provide better accuracies compared to existing feature selection techniques using existing classifiers.

\par Among the 3515 selected features, number of mRNA, lncRNA, miRNA and other RNA are 2496, 533, 14 and 472 respectively.
% \usepackage{multirow}
\begin{table}[]
\caption{Classification performance of different feature selection techniques for selected feature count, k = 500, 1000, 2000, 3000, 3515 and 4000.}
\label{tab:fslistcompare}
\begin{tabular}{|l|l|l|l|l|l|l|l|l|l|}
\hlinewd{1pt}
\textbf{Features} & \makecell[l] {\textbf{Selection method}$\rightarrow$ \\ \textbf{Classifier} $\downarrow$ } & \textbf{Chi2} & \textbf{UFS} & \textbf{MI} & \textbf{LASSO} & \textbf{ENet} & \textbf{RF} & \textbf{RFE} & \textbf{FSFSP} \\ \hlinewd{1pt}
\multirow{7}{*}{500} & LR & 93.26 & 94.25 & 94.46 & 96.33 & 94.85 & 94.70 & 95.98 & \\ \cline{2-9} 
 & SVM & 93.31 & 94.13 & 94.53 & 96.50 & 94.92 & 94.51 & 95.75 & \\ \cline{2-9} 
 & XGB & 94.84 & 95.39 & 94.80 & 95.61 & 94.35 & 95.15 & 96.25 & \\ \cline{2-9} 
 & KNN & 90.09 & 92.35 & 91.16 & 91.40 & 91.20 & 92.66 & 93.82 & \\ \cline{2-9} 
 & RF & 92.69 & 93.29 & 92.74 & 92.68 & 92.08 & 92.97 & 94.71 & \\ \cline{2-9} 
 & mvEns & 94.27 & 94.80 & 95.04 & 96.76 & 95.35 & 95.05 & 96.34 & \\ \cline{2-9} 
 & avEns & 94.91 & 95.33 & 95.35 & 96.89 & 95.50 & 95.39 & 96.60 & \\ \hlinewd{1pt}
\multirow{7}{*}{1000} & LR & 94.82 & 95.39 & 95.69 & 96.61 & 96.03 & 96.03 & 96.40 & \\ \cline{2-9} 
 & SVM & 94.54 & 95.10 & 95.44 & 96.90 & 96.08 & 95.83 & 96.33 & \\ \cline{2-9} 
 & XGB & 95.55 & 95.58 & 95.57 & 95.92 & 95.15 & 95.59 & 96.33 & \\ \cline{2-9} 
 & KNN & 91.61 & 92.48 & 91.96 & 91.95 & 91.80 & 93.15 & 94.15 & \\ \cline{2-9} 
 & RF & 93.61 & 93.48 & 93.75 & 92.98 & 93.11 & 93.17 & 94.72 & \\ \cline{2-9} 
 & mvEns & 95.22 & 95.54 & 95.85 & 96.96 & 96.36 & 96.13 & 96.67 & \\ \cline{2-9} 
 & avEns & 95.58 & 95.91 & 95.90 & 96.99 & 96.32 & 96.47 & 96.81 & \\ \hlinewd{1pt}
\multirow{7}{*}{2000} & LR & 95.98 & 96.19 & 96.23 & 96.46 & 96.51 & 96.31 & 96.65 & \\ \cline{2-9} 
 & SVM & 95.69 & 95.73 & 95.94 & 96.99 & 96.37 & 96.04 & 96.53 & \\ \cline{2-9} 
 & XGB & 96.11 & 95.64 & 95.88 & 95.99 & 95.25 & 95.83 & 96.19 & \\ \cline{2-9} 
 & KNN & 92.56 & 92.89 & 92.53 & 91.65 & 91.77 & 92.94 & 93.96 & \\ \cline{2-9} 
 & RF & 94.12 & 93.66 & 94.13 & 93.41 & 93.30 & 93.94 & 94.50 & \\ \cline{2-9} 
 & mvEns & 96.09 & 96.20 & 96.23 & 96.95 & 96.57 & 96.38 & 96.73 & \\ \cline{2-9} 
 & avEns & 96.28 & 96.35 & 96.51 & 96.97 & 96.51 & 96.57 & 96.80 & \\ \hlinewd{1pt}
\multirow{7}{*}{3000} & LR & 96.39 & 96.36 & 96.71 & 96.41 & 96.68 & 96.53 & 96.78 & \\ \cline{2-9} 
 & SVM & 96.16 & 96.14 & 96.24 & 96.99 & 96.55 & 96.17 & 96.59 & \\ \cline{2-9} 
 & XGB & 96.08 & 95.97 & 95.90 & 96.28 & 95.60 & 95.55 & 96.18 & \\ \cline{2-9} 
 & KNN & 92.98 & 93.12 & 92.90 & 92.01 & 91.82 & 92.99 & 93.68 & \\ \cline{2-9} 
 & RF & 94.36 & 93.92 & 94.27 & 93.41 & 93.42 & 93.93 & 94.61 & \\ \cline{2-9} 
 & mvEns & 96.37 & 96.42 & 96.65 & 96.94 & 96.73 & 96.47 & 96.77 & \\ \cline{2-9} 
 & avEns & 96.53 & 96.68 & 96.87 & 96.96 & 96.75 & 96.56 & 96.82 & \\ \hlinewd{1pt}

\multirow{7}{*}{3515}
 & LR & 96.50 & 96.59 & 96.74 & 96.63 & 96.70 & 96.57 & 96.77 & 96.85 \\ \cline{2-10}
 & SVM & 96.25 & 96.22 & 96.29 & 96.58 & 96.52 & 96.24 & 96.50 & 96.59 \\ \cline{2-10}
 & XGB & 95.96 & 95.97 & 96.04 & 95.67 & 95.58 & 95.84 & 96.18 & 96.75 \\ \cline{2-10}
 & KNN & 93.12 & 93.18 & 93.04 & 91.97 & 92.09 & 92.79 & 93.50 & 94.33 \\ \cline{2-10}
 & RF & 94.33 & 93.99 & 94.29 & 93.67 & 93.59 & 93.80 & 94.53 & 94.94 \\ \cline{2-10}
 & mvEns & 96.40 & 96.56 & 96.59 & 96.80 & 96.78 & 96.56 & 96.73 & 96.88 \\ \cline{2-10}
 & avEns & 96.53 & 96.72 & 96.75 & 96.76 & 96.79 & 96.77 & 96.88 & 97.11 \\ \hlinewd{1pt}

\multirow{7}{*}{4000} & LR & 96.55 & 96.52 & 96.71 & 96.49 & 96.74 & 96.34 & 96.82 & \\ \cline{2-9}
 & SVM & 96.25 & 96.23 & 96.35 & 97.03 & 96.58 & 96.10 & 96.50 & \\ \cline{2-9}
 & XGB & 96.09 & 95.91 & 96.00 & 96.26 & 95.64 & 95.74 & 96.19 & \\ \cline{2-9}
 & KNN & 93.13 & 93.12 & 93.18 & 92.14 & 92.09 & 92.55 & 93.55 & \\ \cline{2-9}
 & RF & 94.38 & 93.98 & 94.33 & 93.60 & 93.54 & 93.96 & 94.43 & \\ \cline{2-9}
 & mvEns & 96.50 & 96.46 & 96.67 & 96.95 & 96.80 & 96.48 & 96.79 & \\ \cline{2-9}
 & avEns & 96.73 & 96.73 & 96.77 & 96.97 & 96.76 & 96.58 & 96.89 & \\ \hlinewd{1pt}
\end{tabular}
\end{table}

\subsection{Classification Results}
\par In this section, we present and discuss the results of the proposed classification methods to classify 33 tumor samples using twelve selected feature sets.

\par The classification performance of various standard ML models and our proposed models (mvEns and avEns), in terms of accuracy, increased significantly for these 12 ranked feature sets compared to the baseline features. \autoref{tab:table33features} reported the comparison in detail. These data indicate that our feature selection method identified informative features that play a crucial role in the classification of pan-cancer gene expression data. Rank10 feature set with 3515 features provides the highest accuracy (97.11\%) with the average voting ensemble (avEns) model.

% \usepackage{multirow}
\begin{table*}[h!]
% \usepackage{multirow}
\caption{Predictive accuracy of different ML models using ranked feature sets for the classification of 33 cancer samples from TCGA gene expression data.}
\label{tab:table33features}
\setlength{\tabcolsep}{0.5em}
\begin{tabular}{|p{1.0cm}|p{.7cm}|p{.95cm}|p{.95cm}|p{.95cm}|p{.95cm}|p{.95cm}|p{.95cm}|p{.95cm}|p{.95cm}|p{.95cm}|p{.95cm}|p{.95cm}|p{.95cm}|}
\hline
\multirow{2}{*}{} & \multicolumn{13}{l|}{TCGA33Tumors: Prediction accuracy (\%) using ranked feature sets} \\ \cline{2-14} 
 &   Base &  Rank12 & Rank11 & Rank10 & Rank9 & Rank8 & Rank7 &
  Rank6 & Rank5 & Rank4 & Rank3 & Rank2 & Rank1 \\ 
\cline{2-14} 
Method &   36017 &   1290 &   2462 &   3515 &   4088 &   4401 &   4699 &   5010 &   5346 &   5688 &   6076 &   6545 &   7170 \\ \hline
LR &   96.62 &   96.20 &   96.73 &   96.85 &   96.92 &   96.94 &   96.86 &   96.82 &   96.89 &   96.92 &   96.92 &   96.83 &   96.89 \\ \hline
SVM &   96.44 &   95.93 &   96.37 &   96.59 &   96.66 &   96.71 &   96.63 &   96.62 &   96.64 &   96.62 &   96.64 &   96.66 &   96.71 \\ \hline
XGB &   96.43 &   96.04 &   96.55 &   96.75 &   96.69 &   96.59 &   96.59 &   96.60 &   96.53 &   96.61 &   96.42 &   96.57 &   96.45 \\ \hline
KNN &   92.39 &   93.54 &   93.97 &   94.33 &   94.22 &   94.21 &   94.28 &   94.21 &   94.18 &   94.14 &   94.19 &   94.12 &   94.07 \\ \hline
RF &   94.18 &   94.01 &   94.76 &   94.94 &   95.02 &   94.81 &   95.02 &   94.97 &   94.91 &   94.81 &   94.87 &   94.78 &   94.79 \\ \hline
mvEns &   96.68 &   96.32 &   96.65 &   96.88 &   96.88 &   96.89 &   96.84 &   96.84 &   96.91 &   96.88 &   96.88 &   96.83 &   96.90 \\ \hline
avEns &   96.87 &   96.50 &   96.95 &  \textbf{97.11} &   97.09 &   97.04 &   97.09 &   97.01 &   97.01 &   97.02 &   97.04 &   97.07 &   97.08 \\ \hline
\end{tabular}
\end{table*}

\subsubsection*{Accuracy Measures} The classification results, in terms of accuracy, are shown in  \autoref{tab:table33features}. The accuracy was calculated as an average of 10-fold cross-validation. The proposed ensemble classifier avEns achieved the best accuracy of 97.11\% for Rank10 feature set (3515 features). For the same feature set accuracy achieved by models LR, SVM, XGBoost, KNN, RF and mvEns are  96.85\%, 96.59\%, 96.75\%, 94.33\%, 94.94\% and 96.88\% respectively. Among all the ranking feature sets best accuracies achieved by LR, SVM, XGBoost, KNN, RF and mvEns are 96.94\% for Rank8 feature set (4401 features), 96.71\% for Rank8 feature set (4401 features), 96.75\% for Rank10 feature set (3515 features), 94.33\% for Rank10 feature set (3515 features), 95.02\% for Rank9 feature set (4088 features) and 96.91\% for Rank5 feature set (5346 features), respectively.

%and mvEns model achieved its best accuracy at 96.91\% for the Rank5 feature set (5346 features).

\subsubsection*{Precision, Recall, F1, and Kappa Scores}
\par To acquire further insight in relation to the classification approaches, we evaluate precision, recall, f1-score, and kappa score from experimental data for TCGA33 dataset and present those metrics in  \autoref{tab:table33compare}. Data in the table confirm that the proposed avEns model performs better to classify the 33 tumor samples compared to other classifiers in terms of precision, recall, f1 score, and kappa score. The evaluated metrics also confirm that the proposed mvEns model performs second best after avEns model to classify 33 tumor samples.

\begin{table}[h!]
\caption{Comparison of performance measures of various machine learning classification models using 3515 selected features.}
\label{tab:table33compare}
\begin{tabular}{|l|c|c|c|c|c|}
\hline
 & \multicolumn{5}{c|}{\textbf{Performance Measures (\%)}} \\ \cline{2-6} 
\textbf{Methods} & \textbf{Accuracy} & \textbf{Precision} & \textbf{Recall} & \textbf{F1-score} & \textbf{Kappa} \\ \hline
LR & 96.85 & 96.94 & 95.37 & 96.83 & 96.70 \\ \hline
SVM & 96.59 & 96.69 & 95.09 & 96.57 & 96.42 \\ \hline
XGB & 96.75 & 96.81 & 94.55 & 96.67 & 96.58 \\ \hline
KNN & 94.33 & 94.49 & 92.44 & 94.23 & 94.05 \\ \hline
RF & 94.90 & 95.44 & 91.73 & 94.35 & 94.64 \\ \hline
mvEns & 96.88 & 96.97 & 95.33 & 96.85 & 96.73 \\ \hline
avEns & 97.11 & 97.16 & 95.48 & 97.07 & 96.97 \\ \hline

\end{tabular}
\end{table}

\par There is no doubt that the average voting ensemble model(avEns) does a better job for classifying TCGA33 dataset using the Rank10 feature set. Moreover, \autoref{tab:table33features} shows that the proposed ensemble models provide higher accuracy over any standalone ML models. That is because an ensemble model combines the predictive power of its component models. Furthermore, the proposed ensemble model avEns provides a higher accuracy than the ensemble model mvEns for all 10 sets of ranking features. For a test sample, model mvEns predicts a class by using the maximum vote of sub-model predictions. However, model avEns predicts a class by averaging sub-models' prediction probabilities. Thus, information loss is higher in mvEns than avEns. Hence, the model avEns provides better accuracy than the model mvEns. Because of the same reason, avEns shows better performance in terms of precision, recall, f1-score, and kappa score.

\subsubsection*{AUC value}
\par
\begin{figure}[htp!]
    \includegraphics[width=17.5cm]{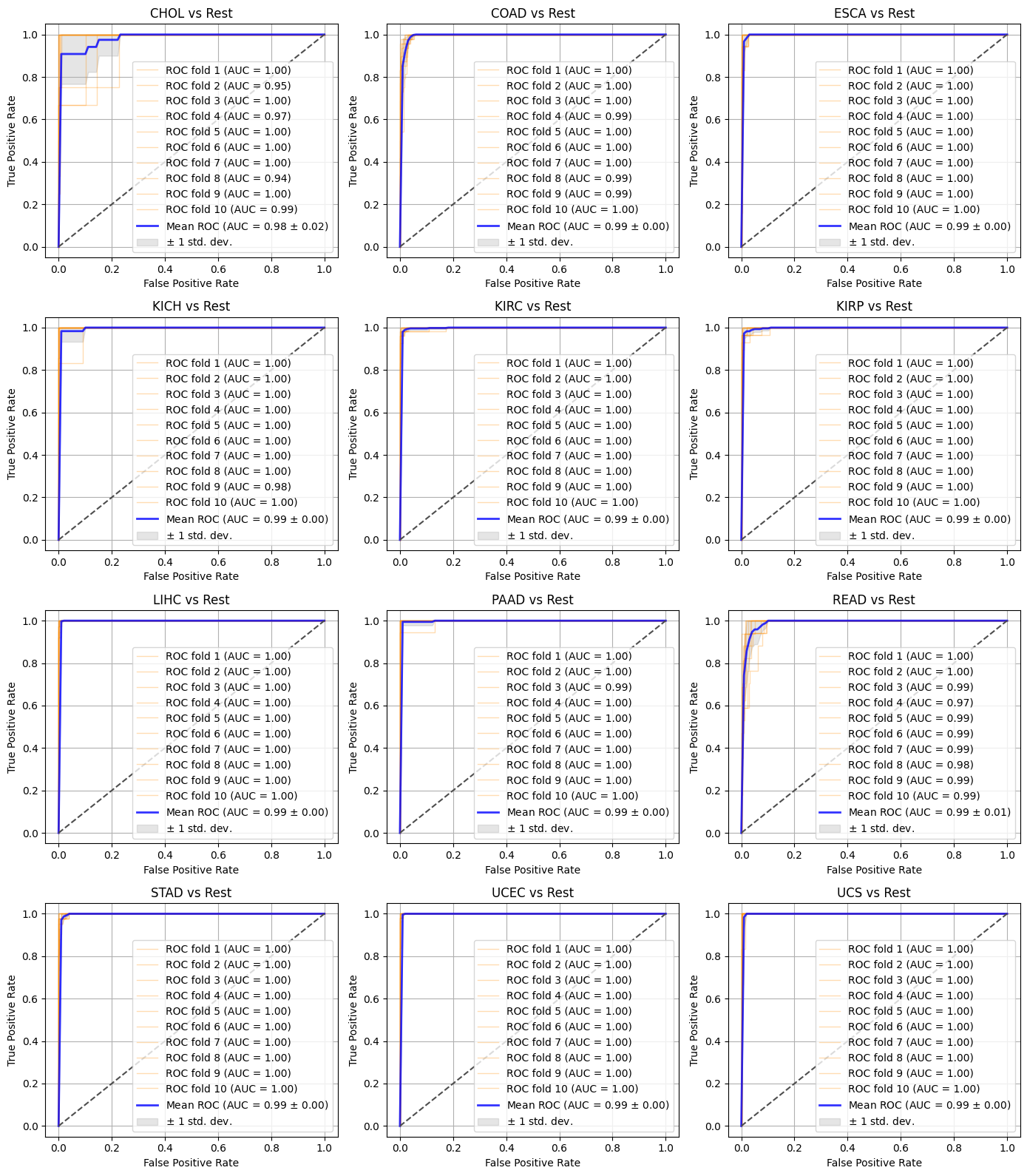}
    \caption{Area under the ROC curve for classification performance of 12 types of tumor with 10-fold cross validation by the average voting ensemble (avEns) model. Classifiers generally struggle to identify these 12 tumors accurately due to the similar tissue of origin issue.}
    \label{fig:aucs}
\end{figure}

 Since our working dataset is highly imbalanced, we calculated the area under the ROC curve for further evaluation. From the result of the avEns classifier model, our achieved macro ROC\_AUC and weighted ROC\_AUC are 0.9993 and 0.9996, respectively. We calculate the AUC value by averaging the AUC values derived from 10-fold cross-validation. In addition to this, we have plotted the mean ROC curves (\autoref{fig:aucs}) for 12 types of tumor, which are critical to classify due to similar tissue of origin factor. To plot the ROC curve, we have used the one-vs-rest (ovr) model within 12 types of tumors. The AUC values are higher than the overall and class-wise accuracies of the classifier. This situation arises when the classifier performs better in identifying true positive cases over true negative cases. 

% The ROC curve is biased towards the positive class. The situation with high AUC and low accuracy can occur when your classifier achieves the good performance on the positive class (high AUC), at the cost of a high false negatives rate (or a low number of true negatives).

\begin{figure*}[h!]
    \includegraphics[width=18cm]{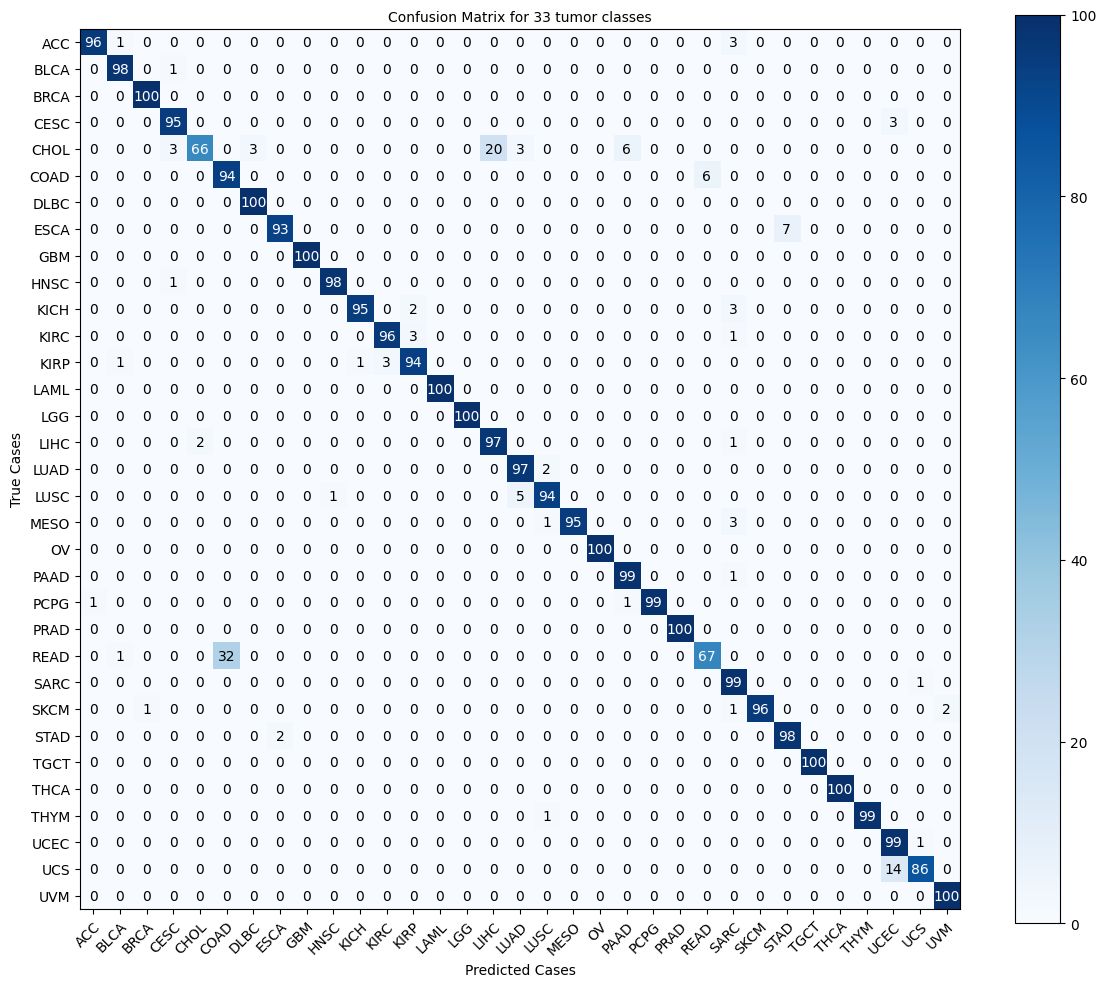}
    \caption{A confusion matrix showing class-wise performance by the proposed averaging ensemble model (avEns) using 3515 features using the Rank10 selected feature set. The confusion matrix was produced by averaging the data from 10-fold cross-validation for 33 tumor classes.}
    \label{fig:33cfm}
\end{figure*}

\subsubsection*{Confusion Matrix for wining avEns classifier on the TCGA33Tumors dataset}

\par The confusion matrix evaluated from the experimental data from the avEns model on the TCGA33Tumors dataset is shown in \autoref{fig:33cfm}, which represents the class-wise prediction summary by the model. The confusion matrix reveals that for 18 tumor types (BLCA, BRCA, DLBC, GBM, HNSC, LAML, LGG, LUAD, OV, PAAD, PCPG, PRAD, SARC, TGCT, THCA, THYM, UCEC, UVM), the model predicted over 97\% accuracy for each class. Three tumor classes, CHOL, READ, and UCS, were predicted with accuracy below 90\% (71\%, 68\%, and 84\% respectively). Moreover, 17\% of CHOL samples and 6\% of PAAD samples were wrongly classified as LIHC, 32\% of rectal adenocarcinoma (READ) samples were wrongly classified as COAD, 14\% of UCS samples were wrongly classified as UCEC and 9\% of ESCA samples were wrongly classified as STAD. These data confirm the existing study\cite{review1,review2,review3,review4,review5,review8,review10} that tumors with the same tissue of origin are classified incorrectly with each other at a higher rate. 

% \usepackage{multirow}
\begin{table*}[h!]
\caption{Performance comparison of the proposed method with existing ones for classifying twelve tumors critical to classifying correctly.}
\label{tab:compare_critial}
\begin{tabular}{|l|l|c|c|c|c|c|c|c|c|}
\hline
\multirow{4}{*}{} & \multirow{2}{*}{} & \multicolumn{8}{l|}{Classification performance (\%)} \\ \cline{2-10} 
 &  & \multicolumn{7}{l|}{Performance in existing literature} & \multicolumn{1}{l|}{This study with} \\ \cline{2-10} 
Sl. & Ref.$\rightarrow$ & \cite{review7} & \cite{review10} & \cite{review1} & \cite{review2} & \cite{review3} & \cite{review4} & \cite{review5} & avEns model \\ \cline{2-10} 
 & No. of Class$\rightarrow$ & (33) & (33) & (31) & (33) & (34) & (34) & (34) & (33)\\ \hline
1 & CHOL & 75 & 56 & 73 & 56 & 57 & 78 & 64 & 66 \\ \hline
2 & COAD & 95 & 95 & 99 & 77 & 88 & 91 & 92 & 94 \\ \hline
3 & ESCA & 85 & 77 & - & 95 & 85 & 85 & 80 & 93  \\ \hline
4 & KICH & 89 & 87 & 96 & 87 & 88 & 86 & 94 & 95 \\ \hline
5 & KIRC & 94 & 95 & 96 & 95 & 97 & 96 & 94 & 96 \\ \hline
6 & KIRP & 94 & 93 & 92 & 93 & 91 & 93 & 91 & 94 \\ \hline
7 & LIHC & 97 & 97 & 98 & 97 & 94 & 95 & 95 & 97 \\ \hline
8 & PAAD & 100 & 97 & 95 & 97 & 97 & 96 & 93 & 99 \\ \hline
9 & READ & 0 & 35 & 0 & - & 61 & 49 & 31 & 67 \\ \hline
10 & STAD & 96 & 96 & - & 35 & 96 & 93 & 91 & 98  \\ \hline
11 & UCEC & 95 & 96 & 96 & 96 & 99 & 98 & 97 & 99 \\ \hline
12 & UCS & 83 & 81 & 62 & 81 & 100 & 82 & 75 & 86 \\ \hline
 \multicolumn{2}{|l|}{Average}  & 83.58 & 83.75 & 80.70 & 82.64 & 87.75 & 86.83 & 83.08 & 90.33 \\ \hline
% \multicolumn{2}{|l|}{Pan-cancer} & 96.81 & 95.59 & 90 & 95.65 & 96.16 & 95.5 & 92.76 & 97.11 \\ \hline
\end{tabular}
\end{table*}

\subsection{Performance Comparison with Existing Literature}
\par Performance of our proposed ensemble model avEns surpasses the classification performance reported in existing literature with similar study (ref. \autoref{tab:literature_review}). We only considered existing research work in which performance was evaluated by multi-fold cross-validation. Wang et al.\cite{review7} reported the highest 96.81\% accuracy for the 33-class TCGA dataset. The proposed models achieved accuracy of 97.11\% (avEns) and 96.88\% (mvEns) for the TCGA33Tumors dataset using the Rank10 feature set with 3515 features (see \autoref{tab:table33features}).

\par We also evaluated our models by comparing the classification performance of 12 critical tumors (CHOL, COAD, ESCA, KICH, KIRC, KIRP, LIHC, PAAD,  READ, STAD, UCEC and UCS) with that of existing work. These critical tumors are prone to being classified incorrectly because each tumor in the set has at least one counterpart within the set that has the same tissue of origin. The comparison statement, shown in \autoref{tab:compare_critial}, depicts that the average classification accuracy of the avEns model for 12 critical tumor classes is 90.33\%, which outperforms all existing works \cite{review7, review10, review1, review2, review3, review4,review5} to the best of our knowledge. The comparison statement also reveals that our approach provides steady-state performance in classifying all 12 types of tumors.

\subsection{GO Pathway Enrichment Analysis}
To further validate the proposed feature selection approach, we performed a gene set enrichment analysis using selected protein-coding mRNAs by ShinyGo\cite{ShinyGO} tool. We used ShinyGO version 0.81 and the KEGG\cite{KEGG1} pathway database with the FDR cut-off value 0.05. The identified top enriched pathways are presented in \autoref{tab:shinygoKEGG} according to the decreasing order of the fold enrichment value.  

\begin{table*}[htbp]
    \caption{Top ten enriched KEGG pathway identified by ShinyGO using selected protein-coding mRNAs.}
    \label{tab:shinygoKEGG}
    \begin{tabular}{|c|l|l|c|c|c|}
        \hline
        \textbf{Sl} & \textbf{Pathway} & \textbf{Title} & \textbf{nGenes} & \textbf{Pathway Genes} & \textbf{Fold Enrichment} \\
        \hline
        1 & hsa04950 & Maturity Onset Diabetes of the Young (MODY) & 14 & 26 & 4.94 \\ \hline
        2 & hsa04610 & Complement and Coagulation Cascades (CCC) & 33 & 85 & 3.56 \\ \hline
        3 & hsa04512 & ECM-receptor interaction (ECM) & 28 & 88 & 2.92 \\ \hline
        4 & hsa04979 & Cholesterol Metabolism & 15 & 51 & 2.70 \\ \hline
        5 & hsa04918 & Thyroid Hormone Synthesis (THS) & 21 & 75 & 2.57 \\ \hline
        6 & hsa04927 & Cortisol Synthesis and Secretion (CSS) & 18 & 65 & 2.54 \\ \hline
        7 & hsa04974 & Protein Digestion and Absorption (PDA) & 28 & 103 & 2.50 \\ \hline
        8 & hsa04520 & Adherens Junction (AJ) & 19 & 71 & 2.45 \\ \hline
        9 & hsa05146 & Amoebiasis & 24 & 102 & 2.16 \\ \hline
        10 & hsa04510 & Focal Adhesion (FA) & 46 & 200 & 2.11 \\ \hline
    \end{tabular}
\end{table*}

\par When the MODY pathway is activated, the patient gets diabetes early, usually before age 30. This makes them more likely than most to get liver and pancreatic tumors\cite{mody1,mody2}. By encouraging the suppression of tumor immunity, the CCC pathway enrichment can favor the growth and spread of tumors\cite{CCC}. The extracellular matrix (ECM) pathway plays a significant structural role in the development of cancer and the tumor microenvironment\cite{ECM}. Tumor growth and progression have been associated with abnormal activity in the cholesterol metabolism\cite{CholMeta} pathway. Thyroid hormones are key regulators of essential cellular processes, including proliferation, differentiation, apoptosis, and metabolism\cite{THS}. Hence, changes in the THS pathway may lead to tumor development in the body. CSS pathway is associated with 23.6\% of cancer patients, especially breast, colorectal, prostate and thyroid cancer\cite{CSS}. The prevalence of malnutrition is high in patients with cancer\cite{PDA}, which leads to eventual muscle and weight loss. The protein digestion and absorption (PDA) pathway abnormality is the factor behind this. Adherens junctions (AJs) pathway controls cell dynamics and tissue integrity\cite{adjunc}. Abnormality in this pathway disrupts a cell's normal regulation and causes it to develop tumors. Entamoeba histolytica, a parasitic amoebozoan, predominantly infects humans and other primates, causing amoebiasis or amoebic dysentery. Chronic infestation of this parasite may lead to a mass of granulation tissue simulating colon cancer\cite{Amoebiasis}. Focal adhesions are large, dynamic protein complexes that build up tissue by connecting the cell's cytoskeleton to the ECM. Cancer cells exhibit highly altered focal adhesion dynamics\cite{FA}. Cancer patients thus experience altered focal adhesion pathways. This discussion reveals that the selected features most significantly enrich pathways related to cancer development.
%\subsection{Limitations and future directions}

\section{Conclusion}
\par In this study, we first present a new approach for selecting informative features for high-dimensional gene expression dataset. Then, for pan-cancer classification, we propose and implement two ensemble classifier models, avEns and mvEns. Based on the tests and results, the winning model (avEns) achieves 97.11\% accuracy. The area under the ROC curve value for the winning model for the dataset is close to 1. These results surpass all previous classification performances on the TCGA 33-class  transcriptome datasets with multi-fold cross-validation, to the best of our knowledge. The model avEns also works better compared to existing methods to classify 12 types of tumors, namely CHOL, LIHC, PAAD, ESCA, STAD, COAD, READ, UCEC, UCS, KICH, KIRC, and KIRP, which are critical to classify with higher accuracy due to the similar tissue of origin issue. The methods we presented in this research work are not limited to the feature selection and classification of pan-cancer gene expression data but are also applicable to any binary (tumor vs. normal) or subtype classification of a cancer.

%\appendices
%\section{\break Footnotes}
%Number footnotes separately in superscript numbers.\footnote{It is recommended that footnotes be %avoided (except for
%the unnumbered footnote with the receipt date on the first page). Instead,
%try to integrate the footnote information into the text.} Place the actual
%footnote at the bottom of the column in which it is cited; do not put
%footnotes in the reference list (endnotes). Use letters for table footnotes
%(see Table \ref{table33compare}).

\section*{Acknowledgments}
The results achieved by this study and shown here are based on data acquired from the \href{https://www.cancer.gov/tcga}{TCGA Research Network}.

\section*{CRediT Author Contributions}
\textbf{Tareque Mohmud Chowdhury:} Conceptualization, Investigation, Data curation, Formal analysis, Methodology, Validation, Writing – Original Draft, Writing – Review and Editing. \textbf{Farzana Tabassum and Sabrina Islam:} Methodology, Validation, Writing – Review and Editing. \textbf{Abu Raihan Mostofa Kamal:} Validation, Writing – Review and Editing, Supervision. 

%All authors have read and approved the final version of the manuscript.

\section*{Competing interests}
The authors declare that they have no known competing financial interests or personal relationships that could have appeared to influence the work reported in this paper.

\section*{Funding Sources}
This research received no specific grant from any funding agency in the public, commercial, or not-for-profit sectors.

\section*{Data and Code Availability Statement}
The data that support the findings of this study are openly available at Harvard Dataverse\cite{tcga33tumors}. The R codes developed to retrieve the gene expression data from the TCGA repository and data cleaning are included in the dataset. Python scripts for feature selection and classification will be made available upon request to the corresponding author. 

%The authors declare no competing interests.

% \bibitem{Abedeen2023}
% \bibinfo{author}{Abedeen, Iftekharul; Rahman, Md. Ashiqur; Zohra Prottyasha, Fatema; Ahmed, Tasnim; Mohmud Chowdhury, Tareque; Shatabda, Swakkhar.}
% \newblock \bibinfo{title}{FracAtlas: A Dataset for Fracture Classification, Localization and Segmentation of Musculoskeletal Radiographs. \emph{figshare}},
%   \url{https://doi.org/10.6084/m9.figshare.22363012} (\bibinfo{year}{2023}).

% \noindent LaTeX formats citations and references automatically using the bibliography records in your .bib file, which you can edit via the project menu. Use the cite command for an inline citation, e.g. \cite{Kaufman2020, Figueredo:2009dg, Babichev2002, behringer2014manipulating}. For data citations of datasets uploaded to e.g. \emph{figshare}, please use the \verb|howpublished| option in the bib entry to specify the platform and the link, as in the \verb|Hao:gidmaps:2014| example in the sample bibliography file. For journal articles, DOIs should be included for works in press that do not yet have volume or page numbers. For other journal articles, DOIs should be included uniformly for all articles or not at all. We recommend that you encode all DOIs in your bibtex database as full URLs, e.g. https://doi.org/10.1007/s12110-009-9068-2.

\end{document}